\documentclass[sigconf]{acmart}

\usepackage{booktabs} 

\usepackage{bm}
\usepackage{multirow}
\usepackage{subfigure}
\usepackage{caption}
\usepackage{algorithm, algorithmic}
\usepackage{balance}
\usepackage{lipsum}

\newcommand*{\affaddr}[1]{#1} 
\newcommand*{\affmark}[1][*]{\textsuperscript{#1}}

\newcommand\blfootnote[1]{%
	\begingroup
	\renewcommand\thefootnote{}\footnote{#1}%
	\addtocounter{footnote}{-1}%
	\endgroup
}

\begin{document}
\copyrightyear{2018} 
\acmYear{2018} 
\setcopyright{acmcopyright}
\acmConference[ICMR '18]{2018 International Conference on Multimedia Retrieval}{June 11--14, 2018}{Yokohama, Japan}
\acmBooktitle{ICMR '18: 2018 International Conference on Multimedia Retrieval, June 11--14, 2018, Yokohama, Japan}
\acmPrice{15.00}
\acmDOI{10.1145/3206025.3206030}
\acmISBN{978-1-4503-5046-4/18/06}

\fancyhead{}
	
\author{
	Wenjie Zhang\affmark[1], Junchi Yan\affmark[1,2], Xiangfeng Wang\affmark[1*] and Hongyuan Zha\affmark[1]\\
	\affaddr{\affmark[1]East China Normal University}\\
	\affaddr{\affmark[2] Shanghai Jiao Tong University}\\
}

\title{Deep Extreme Multi-label Learning}

%
%
%

\begin{abstract}
Extreme multi-label learning (XML) or classification has been a practical and important problem since the boom of big data.
The main challenge lies in the exponential label space which involves $2^L$ possible label sets especially when the label dimension $L$ is huge, {\em{e.g.}}, in millions for Wikipedia labels.
This paper is motivated to better explore the label space by originally establishing an explicit label graph.
In the meanwhile, deep learning has been widely studied and used in various classification problems including multi-label classification, however it has not been properly introduced to XML, where the label space can be as large as in millions.
In this paper, we propose a practical deep embedding method for extreme multi-label classification, which harvests the ideas of non-linear embedding and graph priors-based label space modeling simultaneously.
Extensive experiments on public datasets for XML show that our method performs competitive against state-of-the-art result.
\end{abstract}


\maketitle
\blfootnote{Contact author is Xiangfeng Wang.}
\section{Introduction}\label{sec:intro}
The eXtreme Multi-label Learning (XML) addresses the problem of learning a classifier which can automatically tag a data sample with the most relevant subset of labels from a large label set.
For instance, there are more than a million labels ({\em{i.e.}}, categories) on Wikipedia and it is expect to build a classifier that can annotate a new article or a web page with a subset of relevant Wikipedia categories.
However, XML becomes significantly challenging in order to simultaneously deal with massive labels, dimensions and training samples.
Compared with traditional multi-label learning methods \cite{tsoumakas2006multi}, the \emph{extreme} multi-label learning methods even more focus on tackling the problem with both extremely high input feature dimension and label dimension.
It should also be emphasized that multi-label learning is distinct from multi-class classification \cite{wu2004probability} which only aims to predict a single mutually \emph{exclusive} label.
But in contrast, multi-label learning allows for the co-existence of more than one labels for a single data sample.

One straightforward method is to train an independent one-against-all classifier for each label, which is clearly not optimal for multi-label learning because the dependency between class labels should not be leveraged.
Furthermore, for extreme multi-label leaning, this is not practical since it becomes computationally intractable to train a massive number of classifiers, {\em{e.g.}}, one million classifiers.
Although the issue could possibly be ameliorated if the label hierarchy is established, it is usually impracticable to guarantee this structure in many applications \cite{bhatia2015locally}.
The issue also lies in the prediction stage, where all the classifiers need to be evaluated in every testing data sample procedure.
To address all these challenges, state-of-the-art extreme multi-label learning methods have been proposed recently, which in general can be divided into two categories: tree based methods and embedding based methods.

Tree based methods \cite{weston2013label,agrawal2013multi,prabhu2014fastxml} become popular as they enjoy notable accuracy improvement over traditional embedding based methods.
The idea is how to learn the hierarchy from training data.
Usually the root is initialized to contain the whole label set.
Further the node partition scheme is introduced to determine which labels should be assigned to the left child or the right.
Nodes are recursively partitioned till each leaf contains a small number of labels.
As follows, in the prediction stage, the testing sample passes down the well-established tree until it arrives at the leaf node whereby its labels are finally determined.

\begin{figure*}[tb!]
	\centering
	\subfigure{\includegraphics[width=0.76\textwidth]{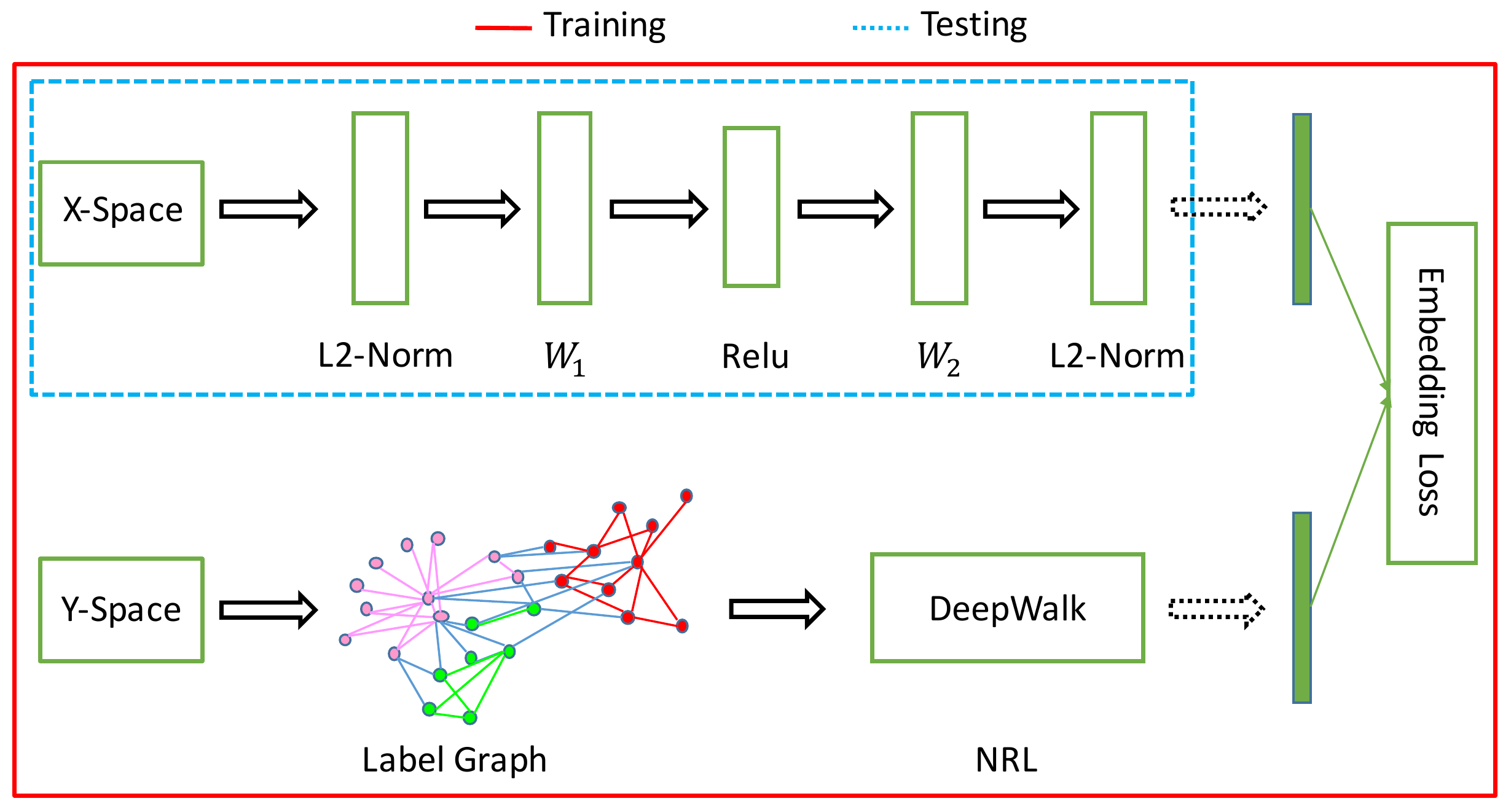}}
	\vspace{-9pt}
	\caption{Approach framework: the solid red box represents the training stage, while the dashed blue box represents the mapping function from the high-dimensional feature space to the low-dimensional non-linear embedding space.
		``L2-Norm" denotes the $\ell_2$-normalization layer, ``$W_{1}$" and ``$W_{2}$" both denote the fully-connected layers, ``Relu" denotes the non-linear activation function,  ``Label Graph" is built from label space (there exist an edge between labels if the two labels ever co-exist with each other in any sample), ``NRL" stands for network representation learning.}
	\label{fig:framework}
\end{figure*}

The embedding based method is another fundamental technique to handle the issue \cite{HsuNIPS09,zhang2011multi,tai2012multilabel,balasubramanian2012landmark,bi2013efficient,cisse2013robust}.
These approaches attempt to make the training and prediction procedures tractable by assuming low-rank training label matrix (of which each column/row corresponds a training sample's label vector) and linearly embedding the high-dimensional label vectors to low-dimensional label vectors to reduce the effective number of labels.
As for the prediction stage, labels for a novel sample are determined through a lifting decompression matrix, which maps the embedded label vectors back to the original extremely high-dimensional label space.
However, the fundamental problem of the embedding method is the low-rank label space assumption, which is violated in most real world applications \cite{bhatia2015locally}.
In general, traditional embedding type methods are usually more computationally efficient than tree based methods but with the expense of accuracy.

Notably, the state-of-the-art embedding based method SLEEC (\textbf{S}parse \textbf{L}ocal \textbf{E}mbeddings for \textbf{E}xtreme Multi-label \textbf{C}lassification) \cite{bhatia2015locally} achieves not only significant accuracy but also computationally efficiency, which is a formulation for learning a small ensemble of local distance preserving embeddings.
It is intriguing to take a deep neural network approach for non-linearity modeling.
To the best of our knowledge, there is very few prior art on deep learning for XML.
Furthermore we observe that the label structure is also very important in tree based methods whereby the labels are denoted by nodes.
However, they seem to be totally ignored in state-of-the-art embedding based methods. For instance, SLEEC focuses on dimension reduction on the raw \emph{label matrix} (of which each column/row corresponds a training sample's label vector) rather than explore label graph structure.

In this paper, we propose a deep learning based approach for extreme multi-label learning.
We extend traditional deep learning framework on multi-label classification problem
by establishing non-linear embedding in both feature and label spaces.
To our best knowledge, our method is devotes an initial attempt to model the feature space non-linearity and label graph structure simultaneously for the XML problem.

The present work possesses several unique contributions compared with previous work:
\begin{itemize}
\item To the best of our knowledge, this is the first work to introduce explicit label graph structure into the \emph{extreme} multi-label learning.
It is worth noting that label graph structure is different from the label hierarchy explored by the tree based methods.
\item An early work for adapting deep learning to the XML setting, and specifically the first work to our best knowledge for using deep network for embedding in XML -- see the approach overview in Figure \ref{fig:framework}. As a matter of fact, we are unable to identify any prior art for exploring deep learning either for feature space reduction or for label space reduction in the XML setting.
\item Extensive experiments on various public benchmark datasets show that without ensemble, our method can perform competitively against or even outperform state-of-the-art ensemble method SLEEC \cite{bhatia2015locally}.

\end{itemize}
The paper is organized as follows.
Section \ref{sec:related} introduces the related work.
The details of the above mentioned concept and idea will be detailed in Section \ref{sec:main}.
Section \ref{sec:implent} and Section \ref{sec:exp} contain implementation details and numerical experiment results respectively, while the Section \ref{sec:con} concludes this paper.

\section{Related Work}\label{sec:related}

\subsection{Classification Methods for XML}
The XML problem is in general algorithmically intractable for one-against-all classifier\footnote{The recent effort \cite{babbar2017dismec} shows that one-vs-rest mechanism is attainable with competitive accuracy against state-of-the-arts {\em{e.g.}} FastXML \cite{prabhu2014fastxml} and SLEEC \cite{bhatia2015locally}, by intensive system level parallel optimization.}, as mentioned in Section \ref{sec:intro}, various tree based and embedding based models are devised.

\smallskip
\noindent \textbf{Tree based methods for XML}:
The label partitioning by sub-linear ranking (LPSR) method \cite{weston2013label} focuses on reducing the prediction time by establishing a hierarchy structure over a benchmark classifier.
However, expensive cost still be an issue since LPSR needs to learn the hierarchy additionally.
The multi-label random forest method (MLRF) \cite{agrawal2013multi} seeks to learn an ensemble of randomized trees instead of relying on a fundamental classifier.
MLRF also suffers the extremely high training cost as LPSR.
FastXML \cite{prabhu2014fastxml}, which is considered the state-of-the-art tree based method for XML, is proposed to learn the hierarchy not over the label space but the feature space. FastXML defines the set of labels active in a region as the union of the labels of all training points in that region.
Predictions are made by returning the ranking list of the leaf nodes according to occurring frequency in the ensemble containing the novel point.

\smallskip
\noindent\textbf{Embedding methods for XML}:
In order to more practically implement the training and prediction procedures, embedding based approaches aim to reduce the effective number of labels by assuming low-rank structure.
Specifically, given $n$ training samples $\left\{ \bm{x}_i,\bm{y}_i \right\}$, $\bm{x}_i\in \mathbb{R}^d$ denotes the feature vector and $\bm{y}_i\in \{0,1\}^{L}$ denotes the related label vector, whereby $d$ and $L$ are both supposed to be huge.
Embedding based methods attempt to establish the linear mapping $\bm{U}\in \mathbb{R}^{\ell \times L}$ which maps the label vectors $\bm{y}_i$ to a significantly low-dimensional vector $\hat{\bm{y}}_i\in {\mathbb{R}}^{\ell}$ ($\ell\ll d,L$), {\em{i.e.}},
$$
\hat{\bm{y}}_i=\bm{U}\bm{y}_i.
$$
The testing procedure is to predict the labels of a testing sample $\bar{\bm{x}}\in \mathbb{R}^d$ by lifting the guaranteed low-dimensional vector $\bar{\bm{y}}\in \mathbb{R}^{\ell}$ to the original label space through another linear or non-linear mapping.

In fact, the main difference of existing embedding based models often lies in the compression and decompression techniques for embedding and lifting respectively, {\em{e.g.}}, compressed sensing \cite{HsuNIPS09}, output codes \cite{zhang2011multi}, SVD-type \cite{tai2012multilabel}, landmark labels \cite{balasubramanian2012landmark,bi2013efficient}, Bloom filters \cite{cisse2013robust} and etc.
The recent embedding based method SLEEC \cite{bhatia2015locally} significantly increases the accuracy of embedding based methods.
SLEEC differs from previous embedding based methods from two perspectives:
i) in the training procedure, instead of training a global mapping matrix to a linear low-rank subspace, it incorporates the non-linear neighborhood constraints in the low-dimensional embedding space;
ii) in the prediction procedure, rather than introducing a decompression matrix for dimension lifting, it employs a simple $k$-nearest neighbor ($k$-NN) classifier in the embedding space.
Our method falls into the embedding based approach group, but we take one step further by exploring neural networks to model the label space structure as well as for feature space embedding.

\subsection{Traditional Multi-label Classification}
Multi-label classification (MLC) is a fundamental problem in machine learning area.
Traditionally MLC can be tackled with a moderate number of labels \cite{tsoumakas2006multi}.
This makes it different from the XML problem where it involves millions of or more labels for each data sample.
Previous MLC methods \cite{boutell2004learning} transform the MLC problem either into one or more single-label classification or regression.
Recent approaches \cite{cheng2010bayes,bi2011multi} try to solve the multi-label learning directly.
However, when the number of labels grows rapidly, these methods can easily become computationally unaffordable.
For instance, if tree based models are employed to traditional MLC \cite{zhang2010multi,bi2011multi}, the trees will become giant with huge feature dimension and samples number, which leads to the intractability for training.
There is also a principled generalization for the naive one-against-all method \cite{hariharan2010large} which cost-effectively explores the label priors, which is similar with one-vs-all method lacking algorithmically scalable ability to XML.

Meanwhile, in cross modal retrieval field, deep neural network \cite{wang2016learning,wang2018learning} has been designed for learning the shared embedding space between images and texts.
However, the input feature vectors are usually with small dimension and therefore existing deep embedding frameworks can not be directly applied to the extreme multi-label learning.

\section{Deep Learning for XML}\label{sec:main}

Our approach involves the training stage and the prediction stage.
In the training stage, we map the high-dimensional feature vector and high-dimensional label vector into a common embedding space, and the graphical display of approach framework is proposed in Figure \ref{fig:framework}.
In the prediction stage, a standard $k$-NN based classifier is used in the embedded feature space to determine the final label predictions.
We first briefly introduce three foundation components of our proposed approach framework and further describe the training and prediction stages in details.

\subsection{Preliminaries}
Let $\mathcal{D}=\left\lbrace \left( \bm{x}_{1},\bm{y}_{1} \right),  \ldots, \left( \bm{x}_{n},\bm{y}_{n} \right) \right\rbrace $ be the given training dataset, $ \bm{x}_{i} \in \mathbb{R}^{d} $ be the input feature vector,  $ \bm{y}_{i} \in \left\lbrace 0,1 \right\rbrace ^L $ be the corresponding label vector, and $(\bm{y}_{i})_j=1$ iff the $\mathit{j}$-th label is turned on for $\bm{x}_{i}$.
Let $X = [ \bm{x}_{1},\cdots, \bm{x}_{n} ] $ be the data matrix and $Y = [ \bm{y}_{1},\cdots, \bm{y}_{n} ] $ be the original label matrix.
For any unseen instance $\bm{x}\in \mathbb{R}^d$, the multi-label classifier $h(\bm{x}):\mathbb{R}^{d} \to \left\lbrace 0,1 \right\rbrace ^L$ predicts the proper labels of $\bm{x}$.
Matrix $V\in\mathbb{R}^{\ell \times L}$ denotes the embedded label matrix, which is learned by the deep label graph embedding detailed as follows.

\subsection{Deep Label Graph Embedding}
In our work, we aim at simultaneously taking advantage of non-linear embedding and label structure information, while the state-of-the-art embedding based method SLEEC ignores to model the label structure.
We focus on using label graph to reflect the label structure.
Our approach can efficiently incorporate the label structure cues, which is missing in many popular embedding based methods as it is nontrivial to incorporate even the classical label hierarchy information.

We first establish a label graph.
\textit{There exists an edge if the two labels ever co-exist with each other in any sample.}
If the prior label hierarchy information is guaranteed, we will directly incorporate it instead of establishing the label graph.
Further we intend to learn a representation, where we assume that two nearby labels in the graph have similar representation.
Inspired by word2vec, which learns word representation based on the sentence context information, we exploit the local context information of the labels by exploiting the label node context information in the graph.
There are many advanced methods for network representation leaning (NRL) or network embedding, which studies the problem of embedding information networks into low-dimensional spaces, {\em{i.e.}}, each vertex is represented by a low-dimensional vector.
The precedent work DeepWalk \cite{perozzi2014deepwalk} uses random walk to collect sequences as contexts and establish the word2vec representation for all nodes in the graph.
In this paper, we adopt DeepWalk in our framework, because of its significant computational efficiency and scalability.


\subsection{Deep Non-linear Embedding}
In this paper, we aim to learn a non-linear mapping from $X$ to the embedding space, while previous embedding based methods usually transfer the high-dimensional feature vectors to a low-dimensional space under a linear mapping.
The feature representation of $\bm{x}\in \mathbb{R}^{d}$ in the embedding space is denoted as $f_{\bm{x}} := F (\bm{x};\bm{W})\in \mathbb{R}^{\ell}$ where $\bm{W}$ denotes the embedding parameter, and the feature representation of $\bm{y}\in \mathbb{R}^{L}$ in the latent space is denoted as $f_{\bm{y}}\in \mathbb{R}^{\ell}$.
Therefore, the distance between $\bm{x}$ ({\em{e.g.}} document features) and the corresponding $\bm{y}$ ({\em{e.g.}} label features) can be represented by the distance between $f_{\bm{x}}$ and $f_{\bm{y}}$, {\em{i.e.}},
\begin{equation}
d (\bm{x},\bm{y}):= d (f_{\bm{x}},f_{\bm{y}}) = \sum_{j}^{\ell} h \left( ( f_{\bm{x}} )_{j},  (f_{\bm{y}} )_{j} \right),
\end{equation}where
\begin{equation}
h (a,b)=
\left\{
\begin{array}{ll}
0.5(a - b)^2 & \hbox{if } \left| a - b \right| \leq 1,\\
|a - b|-0.5 & \hbox{otherwise},\\
\end{array}
\right.
\end{equation}which denotes the robust smooth-$\ell_{1}$ loss \cite{girshick2015fast} which is less sensitive to outliers than the classical $\ell_{2}$ loss.
The objective function that we will minimize can be formulated as
\begin{equation}
\mathcal{L}(\bm{W}) := \sum_{i=1}^n d (\bm{x}_i,\bm{y}_i) = \sum_{i=1}^{n}d (f_{\bm{x}_i},f_{\bm{y}_i}),
\end{equation}where $\bm{W}$ represents the model parameters for $X$ branch network.
In Figure \ref{fig:framework}, the embedding loss refers to a class of deep metric learning loss such as contrastive loss \cite{chopra2005learning} or triplet loss \cite{schroff2015facenet,hoffer2015deep}, as well as their variants which are widely used in image retrieval, face recognition and person re-identification.
Although they are able to achieve good performance, but the training stage is still difficult and its overhead is slightly larger, because we need to sample lots of appropriate tuples or triplets.
In this paper, with the purpose to propose a much simpler and more effective method, we use the smooth-$\ell_1$ loss instead in our proposed framework.

\subsection{Cluster Embedding Space}

State-of-the-art embedding based method SLEEC introduces a clustering step of all training points in high-dimensional feature space as pre-processing.
Further they apply their approach in each of the clusters separately for large-scale datasets.
As we know owing to the curse-of-dimensionality, clustering usually turns out to be unstable for large dimensional feature datasets and frequently leads to some drop in the prediction accuracy.
To safeguard against such instability, they use an ensemble of models which are generated with different sets of clusters, although this technique will increase the model size and training cost.
In our paper, instead of clustering all the training points in high-dimensional feature space, we only cluster all the training points in low-dimensional embedding space after our single learner established.
The clustering step becomes significantly more computational efficient and stable.

\begin{algorithm}[tb]
	\caption{DXML: Training Algorithm}\label{train_algo_proposed}
	\begin{algorithmic}[1]
		\REQUIRE $\mathcal{D}=\left\lbrace \left( \bm{x}_{1},\bm{y}_{1} \right),  \ldots, ( \bm{x}_{n},\bm{y}_{n} ) \right\rbrace $, Embedding dimensionality: $\ell$, No. of clusters: $m$, No. of iteration: $T$
		\STATE Build the label graph, in which there exist an edge between labels if the two labels ever co-exist with each other in any sample;
		\STATE Use DeepWalk~\cite{perozzi2014deepwalk} transfer label graph to embedded label matrix $\bm{V}$;
		\STATE Project original label matrix $Y$ to
		$$f_{Y} = \left\{ f_{\bm{y}}\ \big|\ f_{\bm{y}} = \frac{1}{nnz (\bm{y})} \bm{V} \bm{y},\ \bm{y}\in Y \right\};$$
		\STATE Train the deep neural network shown in Figure \ref{fig:framework} to obtain the mapping from original feature vector set $X$ to embedded feature vector set $f_{X} = \left\{ f_{\bm{x}} \right\}$. Update $\bm{W}$ for $T$ epochs to
		\begin{displaymath}
		\min \mathcal{L}(\bm{W}) = \sum_{i=1}^{n}d (f_{\bm{x}_i},f_{\bm{y}_i});
		\end{displaymath}
		\STATE Partition $f_{X}$ into $Z^{1}, \ldots, Z^{m}$.
		\ENSURE $\left\lbrace Z^{1}, \ldots, Z^{m} \right\rbrace $
	\end{algorithmic}
	\normalsize
\end{algorithm}

\begin{algorithm}[tb]
	\caption{DXML: Test Algorithm}\label{test_algo_proposed}
	\begin{algorithmic}[1]
		\REQUIRE Test point $\bm{x}$, No. of $k$-NN: $k$, No. of desired labels: $p$
		\STATE $\bar{\bm{x}} \leftarrow f(\bm{x})$
		\STATE $Z^{i^{\star}}$: partition closest to $\bar{\bm{x}}$;
		\STATE $K_{\bar{\bm{x}}} \leftarrow k$ nearest neighbors of $\bar{\bm{x}}$ in $Z^{i^{\star}}$;
		\STATE $P_{\bm{x}} \leftarrow$ empirical label dist, for points $\in K_{\bar{\bm{x}}}$;
		\STATE $\bm{y}_{pred} \leftarrow Top_{p} (P_{\bm{x}})$
		\ENSURE $\bm{y}_{pred}$
		
	\end{algorithmic}
	\normalsize
\end{algorithm}

\subsection{Algorithm Framework}\label{sec:implent}

The training algorithm is shown in Algorithm \ref{train_algo_proposed}, termed by DXML (\textbf{D}eep embedding for \textbf{XML} classification).
The test algorithm is shown in Algorithm \ref{test_algo_proposed}.
In the following, we will discuss more details of these two stages.

\medskip
\noindent{\textbf{Training stage}}
\begin{itemize}
\item Label space embedding:
We first build the label graph, in which each node denotes a label and there is an edge if the two connecting labels co-appear on at least one sample in Figure \ref{fig:framework}.
Then, we use DeepWalk to learn low-dimensional continuous feature representations, which is the embedded label matrix $\bm{V}\in \mathbb{R}^{\ell\times L}$ for all nodes in the label graph.

\item Feature embedding:
Given the low-dimensional representation of $\bm{y}\in\{0,1\}^{L}$ by $f_{\bm{y}} = \frac{1}{nnz( \bm{y} )}\bm{V}\bm{y}$, where $nnz(\bm{y})$ denotes the number of non-zero elements of the original binary vector $\bm{y}$.
The $X$ branch deep neural network is trained to find the mapping that the sample's feature vector is embedded into the same space with the low-dimensional representation of $\bm{y}$.
Moreover $f_{\bm{x}}$ and $f_{\bm{y}}$ are close to each other as measured by the embedding loss.

\item Cluster embedding space:
After getting the feature embedding set $\left\{ f_{\bm{x}_i} \right \}$, we partition the set into several clusters with $k$-means.

\end{itemize}

\noindent{\textbf{Prediction stage}}
The prediction is relatively standard: given a test sample $\bm{x}$, we first get its low-dimensional feature representation $f_{\bm{x}}$, then assign the index for it by
\begin{equation}
i^{\star} = \arg \min_{i\in \left\{ 1,\cdots,m \right\} } \left\| f_{\textbf{x}} - {\bm{z}}_{c}^i \right\|_{2},
\end{equation}where ${\bm{z}}_{c}^i$ denotes the center of the $i$-th cluster $Z^i$.
In the following, we perform $k$-NN search in its low-dimensional feature representation to find its similar samples from the training dataset in the corresponding cluster $Z^{i^{\star}}$.
Finally, the average (or simply sum) of the $k$-nearest neighbors' labels is set as the final label prediction.

\begin{table*}[htb!]
	\centering
	\caption{Statistics of the datasets used in experiments. Point denotes the data sample.}
	\resizebox{0.82\textwidth}{!}{
		\begin{tabular}{rrrrrrrr}
			\toprule
			\textbf{Scale}&\textbf{Dataset} & \textbf{Train} & \textbf{Test} & \textbf{Features} & \textbf{Labels} & \textbf{\begin{tabular}[r]{@{}r@{}}Avg. points\\ per label\end{tabular}} & \textbf{\begin{tabular}[r]{@{}r@{}}Avg. labels\\ per point\end{tabular}} \\ \midrule
			\multirow{4}{*}{Small}
			&Bibtex           & 4,880           & 2,515          & 1,836              & 159             & 111.71                                                                  & 2.40                                                                    \\
			&Delicious        & 12,920          & 3,185          & 500               & 983             & 311.61                                                                  & 19.03                                                                   \\
			&MediaMill        & 30,993          & 12,914         & 120               & 101             & 1902.15                                                                 & 4.38                                                                    \\ \midrule
			\multirow{3}{*}{Large}&Wiki10-31K       & 14,146          & 6,616          & 101,938            & 30,938           & 8.52                                                                    & 18.64                                                                   \\
			&Delicious-200K	   & 196,606         & 100,095        & 782,585            & 205,443          & 72.29                                                                   & 75.54                                                                   \\
			&Amazon-670K       & 490,449        & 153,025        & 135,909          & 670,091          & 3.99                                                                   & 5.45                                                                    \\ \bottomrule
	\end{tabular}}
	\label{tab:statistics}
\end{table*}

\begin{table*}[htb!]
	\centering
	\caption{P@k on small scale datasets.}
	\resizebox{0.82\textwidth}{!}{
		\begin{tabular}{rr|rrrrrrr|rr|rr}
			\toprule
			\multirow{2}{*}{Dataset}                   & \multirow{2}{*}{P@k}                 &  & \multicolumn{6}{c|}{Embedding based}              & \multicolumn{2}{c|}{Tree based} & \multicolumn{2}{c}{Other} \\
			& & \textbf{DXML}      & SLEEC & LEML  & WSABIE & CPLST & CS    & ML-CSSP & FastXML         & LPSR         & 1-vs-All      & KNN       \\ \midrule
			\multirow{3}{*}{Bibtex}    & P@1                  & \textbf{66.03}      & 65.08 & 62.54 & 54.78  & 62.38 & 58.87 & 44.98   & 63.42           & 62.11        & 62.62         & 57.04     \\
			& P@3                  & \textbf{40.21 }     & 39.64 & 38.41 & 32.39  & 37.84 & 33.53 & 30.43   & 39.23           & 36.65        & 39.09         & 34.38     \\
			& P@5                  & 27.51      & \textbf{28.87} & 28.21 & 23.98  & 27.62 & 23.72 & 23.53   & 28.86           & 26.53        & 28.79         & 25.44     \\ \midrule
			\multirow{3}{*}{Delicious} & P@1                  & 68.17      & 67.59 & 65.67 & 64.13  & 65.31 & 61.36 & 63.04   & \textbf{69.61}           & 65.01        & 65.02         & 64.95     \\
			& P@3                  & 62.02      & 61.38 & 60.55 & 58.13  & 59.95 & 56.46 & 56.26   & \textbf{64.12}           & 58.96        & 58.88         & 58.89     \\
			& P@5                  & 57.45      & 56.56 & 56.08 & 53.64  & 55.31 & 52.07 & 50.16   & \textbf{59.27}           & 53.49        & 53.28         & 54.11     \\
			\midrule
			\multirow{3}{*}{MediaMill} & P@1                  & \textbf{88.73}      & 87.82 & 84.01 & 81.29  & 83.35 & 83.82 & 78.95   & 84.22           & 83.57        & 83.57         & 82.97     \\
			& P@3                  & \textbf{74.05}      & 73.45 & 67.20 & 64.74  & 66.18 & 67.32 & 60.93   & 67.33           & 65.78        & 65.50         & 67.91     \\
			& P@5                  & 58.83     & \textbf{59.17} & 52.80 & 49.83  & 51.46 & 52.80 & 44.27   & 53.04           & 49.97        & 48.57         & 54.23     \\
			\bottomrule
	\end{tabular}}
	\label{tab:small_result_P}
\end{table*}

\begin{table*}[htb!]
	\centering
	\caption{P@k on large scale datasets.}
	\resizebox{0.55\textwidth}{!}{
		\begin{tabular}{rrrrr|rr}
			\toprule
			&\multicolumn{4}{c|}{Embedding based}              & \multicolumn{2}{c}{Tree based} \\
			Dataset    &  P@k& \textbf{DXML} & SLEEC & LEML  & FastXML & LPSR-NB \\ \midrule
			\multirow{3}{*}{Wiki10-31K}          & P@1                     & \textbf{86.45} & 85.88 & 73.47 & 83.03   & 72.72   \\
			& P@3                     & 70.88 & \textbf{72.98} & 62.43 & 67.47   & 58.51   \\
			& P@5                     & 61.31 & \textbf{62.70} & 54.35 & 57.76   & 49.50   \\ \midrule
			\multirow{3}{*}{Delicious-200K} & P@1                     & \textbf{48.13} & 47.85 & 40.73 & 43.07   & 18.59   \\
			& P@3                     & \textbf{43.85} & 42.21 & 37.71 & 38.66   & 15.43   \\
			& P@5                     & \textbf{39.83} & 39.43 & 35.84 & 36.19   & 14.07   \\ \midrule
			\multirow{3}{*}{Amazon-670K}       & P@1                     & \textbf{37.67}& 35.05 & 8.13 & 36.99   & 28.65   \\
			& P@3                     & \textbf{33.72} & 31.25 & 6.83 & 33.28   & 24.88   \\
			& P@5                     & 29.86 & 28.56 & 6.03  & \textbf{30.53}   & 22.37   \\ \bottomrule
	\end{tabular}}
	\label{tab:large_result_P}
\end{table*}

\begin{table*}[htb!]
	\centering
	\caption{nDCG@k on small scale datasets.}
	\resizebox{0.82\textwidth}{!}{
		\begin{tabular}{rr|rrrrrrr|rr|rr}
			\toprule
			\multirow{2}{*}{Dataset}                   & \multirow{2}{*}{nDCG@k}                 &  & \multicolumn{6}{c|}{Embedding based}              & \multicolumn{2}{c|}{Tree based} & \multicolumn{2}{c}{Other} \\
			& & \textbf{DXML}      & SLEEC & LEML  & WSABIE & CPLST & CS    & ML-CSSP & FastXML         & LPSR         & 1-vs-All      & KNN       \\ \midrule
			\multirow{3}{*}{Bibtex}    & nDCG@1                  & \textbf{66.03}    & 65.08 & 62.54 & 54.78  & 62.38 & 58.87 & 44.98   & 63.42           & 62.11        & 62.62         & 57.04     \\
			& nDCG@3                  & \textbf{60.71}      & 60.47 & 58.22 & 50.11  & 57.63 & 52.19 & 44.67   & 59.51           & 56.50        & 59.13         & 52.29     \\
			& nDCG@5                  & 62.03      & \textbf{62.64} & 60.53 & 52.39  & 59.71 & 53.25 & 47.97   & 61.70           & 58.23        & 61.58         & 54.64     \\ \midrule
			\multirow{3}{*}{Delicious} & nDCG@1                  & 68.17      & 67.59 & 65.67 & 64.13  & 65.31 & 61.36 & 63.04   & \textbf{69.61}           & 65.01        & 65.02         & 64.95     \\
			& nDCG@3                  & 63.45      & 62.87 & 61.77 & 59.59  & 61.16 & 57.66 & 57.91   & \textbf{65.47}           & 60.45        & 60.43         & 60.32     \\
			& nDCG@5                  & 59.89      & 59.28 & 58.47 & 56.25  & 57.80 & 54.44 & 63.36   & \textbf{61.90 }          & 56.38        & 56.28         & 56.77     \\
			\midrule
			\multirow{3}{*}{MediaMill} & nDCG@1                  & \textbf{88.73}      & 87.82 & 84.01 & 81.29  & 83.35 & 83.82 & 78.95   & 84.22           & 83.57        & 83.57         & 82.97     \\
			& nDCG@3                  & \textbf{81.87 }     & 81.50 & 75.23 & 72.92  & 74.21 & 75.29 & 68.97   & 75.41           & 74.06        & 73.84         & 75.44     \\
			& nDCG@5                  & \textbf{80.03}     & 79.22 & 71.96 & 69.37  & 70.55 & 71.92 & 62.88   & 72.37          & 69.34        & 68.18         & 72.83     \\
			\bottomrule
	\end{tabular}}
	\label{tab:small_result_G}
\end{table*}

\begin{table*}[htb!]
	\centering
	\caption{nDCG@k on large scale datasets.}
	\resizebox{0.55\textwidth}{!}{
		\begin{tabular}{rrrrr|rr}
			\toprule
			&\multicolumn{4}{c|}{Embedding based}              & \multicolumn{2}{c}{Tree based} \\
			Dataset    &  nDCG@k& \textbf{DXML} & SLEEC & LEML  & FastXML & LPSR-NB \\ \midrule
			\multirow{3}{*}{Wiki10-31K}          & nDCG@1                     & \textbf{86.45} & 85.88 & 73.47 & 83.03   & 72.72   \\
			& nDCG@3                     & \textbf{76.32} & 76.02 & 64.92 & 71.01   & 61.71   \\
			& nDCG@5                     & 67.11 & \textbf{68.13} & 58.69 & 63.36   & 54.63   \\ \midrule
			\multirow{3}{*}{Delicious-200K} & nDCG@1                     & \textbf{48.13} & 47.85 & 40.73 & 43.07   & 18.59   \\
			& nDCG@3                     & \textbf{43.95} & 43.52 & 38.44 & 39.70   & 16.17   \\
			& nDCG@5                     & \textbf{41.73} & 41.37 & 37.01 & 37.83   & 15.13   \\ \midrule
			\multirow{3}{*}{Amazon-670K}       & nDCG@1                     & \textbf{37.67}& 35.05 & 8.13 & 36.99   & 28.65   \\
			& nDCG@3                     & 33.93 & 32.74 & 7.30 & \textbf{35.11}   & 26.40   \\
			& nDCG@5                     & 32.47 & 31.53 & 6.85  & \textbf{33.86}   & 25.03   \\ \bottomrule
	\end{tabular}}
	\label{tab:large_result_G}
\end{table*}

\begin{table}[htb!]
	\caption{Comparing training and test times (in seconds) of different methods.}
	\resizebox{0.48\textwidth}{!}{
	\begin{tabular}{cccccc}
		\hline
		&                & \multicolumn{2}{c}{DXML} & \multicolumn{2}{c}{SLEEC} \\
		Scale & Dataset        & train (GPU) & test (CPU) & train(CPU)   & test(CPU)  \\ \hline
		& Bibtex         & 353           & 2.1          &  307          & 1.5          \\
		Small & Delicious      & 1630           & 7.8          & 1763            & 6.1         \\
		& MediaMill      & 9403           & 40.2          & 9906           & 35.3          \\ \hline
		& Wiki10-31K     &  3047          & 16.3          & 3241            & 11.4          \\
		Large & Delicious-200K & 14537           & 1471          & 16302            & 1347          \\
		& Amazon-670K    &  67038          & 3573         & 71835           & 3140         \\ \hline
	\end{tabular}
}
	\label{tab:cost_time}
\end{table}
\section{Experiments}\label{sec:exp}

Experiments are carried out on publicly available XML benchmark datasets from the Extreme Classification Repository\footnote{\tiny http://research.microsoft.com/en-us/um/people/manik/downloads/XC/XMLRepository.html}, which contains both small-scale datasets \cite{prabhu2014fastxml} and large-scale datasets \cite{bhatia2015locally}.
We compare with some state-of-the-art peer methods, which includes both embedding based and tree based methods.

\subsection{Protocol}

\smallskip
\noindent\textbf{Platform} The experiments are implemented under CentOS-6.5 64-bit system, with Intel(R) Xeon(R) CPU E5-2680 v2 @ 2.80GHz $\times$6 CPU, GeForce GTX 1080 Ti and 128G RAM.
The code is implemented parlty in Python and partly in MATLAB and the network is built by MXNet\footnote{\tiny https://github.com/apache/incubator-mxnet} \cite{chen2015mxnet}, which is a flexible and efficient library for deep learning and has been chosen by Amazon as the official deep learning framework for its web service.

\smallskip
\noindent\textbf{Datasets}
The extreme multi-label classification datasets include Amazon-670K \cite{mcauley2013hidden}, Delicious-200K \cite{wetzker2008analyzing} and Wiki10-31K \cite{zubiaga2012enhancing}.
All the datasets are publicly available.
It should be noted that some other methods are not scalable enough on such large datasets.
Furthermore, we also present comparisons on public relatively small datasets such as BibTex \cite{katakis2008multilabel}, Delicious \cite{TsoumakasPKDD08}, MediaMill \cite{snoek2006challenge}.
The statistics of the benchmarks are listed in Table \ref{tab:statistics}.

\smallskip
\noindent\textbf{Baseline algorithms for comparison}
Our primary focus is to compare with those state-of-the-art extreme multi-label classification methods, such as embedding based methods like SLEEC \cite{bhatia2015locally} and LEML \cite{yu2014large}, tree based methods like FastXML \cite{prabhu2014fastxml} and LPSR \cite{weston2013label}.
SLEEC was shown to beat all other embedding based methods on the benchmark datasets and LEML uses trace norm regularization to identify a low-dimensional representation of the original high-dimensional label space.
Our method can be considered as a natural combination of label graph and deep embedding. Techniques such as compressed sensing (CS) \cite{HsuNIPS09}, CPLST \cite{ChenNIPS12}, ML-CSSP \cite{bi2013efficient}, one-vs-all \cite{hariharan2012efficient} can only be trained on small datasets using commodity computational hardware.

\smallskip
\noindent\textbf{Evaluation metrics}
The evaluation metric in \cite{bhatia2015locally} is $Precision@k$ ($P@k$) has been widely adopted as the metric of choice for evaluating multi-label algorithms. The $P@k$ is the fraction of correct positive labels, which is the number of correct predictions over $k$. It decrease with the number of $k$ increases. Such metric encourages the correct label to be ranked higher. $Precision@k$ and $nDCG@k$ defined for a predicted score vector $\hat{\bm{y}}\in{\mathbb{R}^L}$ and ground truth label vector $\bm{y}\in{\left\lbrace 0,1\right\rbrace ^L}$ as
\[ P@k= \frac{1}{k}\sum_{j\in{rank_{k}(\hat{\bm{y}})}}\bm{y}_{j}, \]
\[ DCG@k= \sum_{j\in{rank_{k}(\hat{\bm{y}})}}\frac{\bm{y}_{j}}{\log{(j+1)}}, \]
\[ nDCG@k= \frac{DCG@k}{\sum_{j=1}^{\min(k,\left\| \bm{y}\right\|_{0} )}1/\log{(j+1)}}, \]
where $rank_{k}(\bm{y})$ returns the $k$ largest indices of $\bm{y}$ ranked in descending order.

\smallskip
\noindent\textbf{Network Structure}
We propose to learn a non-linear embedding in a deep neural network framework.
As shown in the top half of Figure \ref{fig:framework}, it composes of a fully connected layer with weight $W_{1}$, a successive Rectified Linear Unit (\textbf{ReLU}) layers and another fully connected layer with weight $W_{2}$.
Set $W_1 \in \mathbb{R}^{d\times 256}$ and $W_2\in \mathbb{R}^{256\times \ell}$ for small-scale datasets, while $W_1 \in \mathbb{R}^{d\times 512}$ and $W_2\in \mathbb{R}^{512\times \ell}$ for large-scale datasets.
We apply dropout technique \cite{srivastava2014dropout} right after the last linear layer.
At the end of the network, we add a $\ell_2$ normalization layer.

\smallskip
\noindent\textbf{Hyper-parameter Setting}
For our method, we set the embedding dimension $\ell$ to be $100$ for small-scale and $300$ for large-scale datasets respectively.
We train our networks using vanilla SGD (stochastic gradient descent) with momentum $0.9$, weight decay $0.0005$ and the fixed learning rate $0.015$.
The remaining key hyper-parameter $k$ in $k$-NN is set by cross-validation on a validation set.

\subsection{Results and Ablation Analysis}
The performance of all methods in $P@k$ and $nDCG@k$ on all six datasets are summarized in Tables  \ref{tab:small_result_P}, \ref{tab:large_result_P}, \ref{tab:small_result_G} and \ref{tab:large_result_G}. In general, {\bf{DXML}} performs competitively even without ensemble.

\smallskip
\noindent\textbf{Results on small datasets} The results in Tables \ref{tab:small_result_P} and \ref{tab:small_result_G} are averaged over $10$ train-test split for each dataset. From two tables, one can see that our method DXML can mostly be ranked as top $2$ on all the three datasets. DXML almost outperforms all the other baseline algorithms. On Bibtex and MediaMill datasets, DXML outperforms LEML and FastXML by nearly $4\%$ for $\left\lbrace P,nDCG \right\rbrace @ \left\lbrace 3,5 \right\rbrace$ and outperforms state-of-the-art embedding based method SLEEC by nearly $1\%$. On Delicious dataset, DXML also outperforms SLEEC by nearly 1\% for  $\left\lbrace P,nDCG \right\rbrace @ \left\lbrace 1,3,5 \right\rbrace$, while slightly underperforms the tree based method FastXML.

\smallskip
\noindent \textbf{Results on large datasets} As observed from Table \ref{tab:large_result_P} and \ref{tab:large_result_G}, DXML's predictions could be significantly more accurate than all the other baseline methods (except on Delicious-200K where DXML was ranked second). On the Amazon-670K dataset, our method outperforms LFML and LPSR-NB by around $28\%$ and $9\%$ respectively for $\left\lbrace P,nDCG \right\rbrace @ \left\lbrace 1,3,5 \right\rbrace$ .

\smallskip
\noindent \textbf{Training and test time} The training and test time in seconds for our deep learning based method DXML and state-of-the-art embedding method SLEEC on the six benchmark datasets are summarized in Table \ref{tab:cost_time}. Notice that  DXML and SLEEC are implemented for running on CPU at test stage, while DXML are implemented for running on GPU and SLEEC are implemented for running on CPU at training stage  separately. From the table, one can see that our method DXML has comparable time complexities as SLEEC in testing.

\begin{figure}[ht!]
	\subfigure{\includegraphics[width=0.3\textwidth]{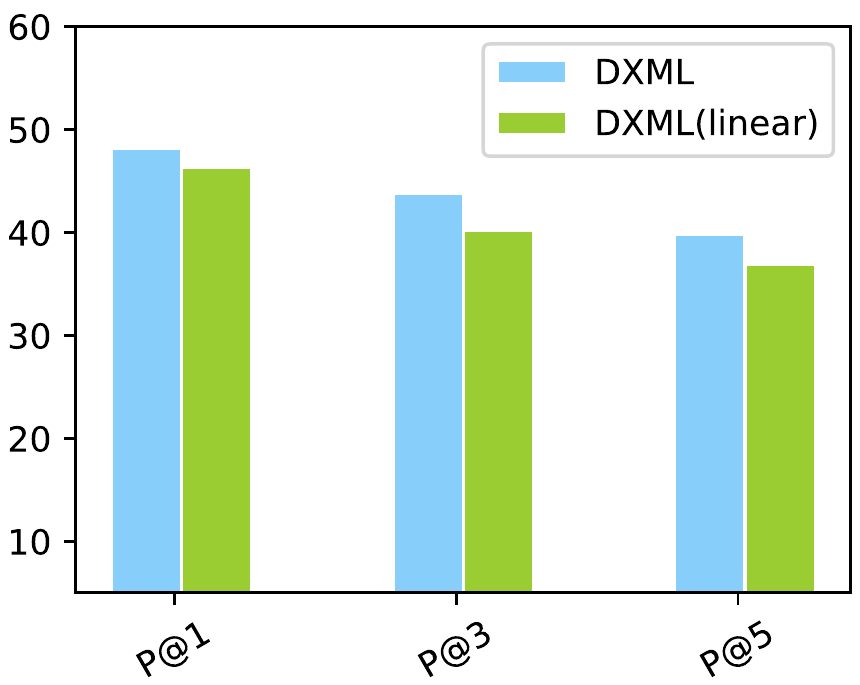}}\vspace{-10pt}
	\caption{Ablation test for precision@$k$ comparison of deep non-linear embedding (see Figure \ref{fig:framework}) and deep linear embedding on  Delicious-200K.}
	\label{fig:ablation}
\end{figure}

\smallskip
\noindent \textbf{Results for ablation test} We also perform ablation test to study the efficacy of non-linear embedding in our framework (see Figure \ref{fig:framework}) compared with the linear embedding, which do not use non-linear activation function (e.g. \textbf{ReLU}) in our framework. As shown in Figure \ref{fig:ablation} DXML with nonlinearities outperforms the DXML without nonlinearities.

In summary, extensive experiments were carried out on six extreme multi-label classification benchmark datasets demonstrating that our deep learning based model DXML can often achieve higher prediction accuracies as compared to the-state-of-the-art embedding method SLEEC. In particular, DXML also shows competitiveness not only on large datasets but also on smaller ones. our method exhibits certain nice properties over SLEEC: i) DXML can be trained with off-the-shelf stochastic optimization solvers and its memory cost can be flexibly adjusted by the mini-batch size. ii) For new arrival data, SLEEC has to be trained from scratch, while in contrast, DXML allows for incremental learning too.

\section{Conclusion}\label{sec:con}
For the extreme multi-label learning (XML) problem, this paper starts with modeling the large-scale label space via the label graph. In contrast, existing XML methods either explore the label hierarchy as done by many tree based method or perform dimension reduction on the original label/sample matrix. Moreover, a deep neural network is devised to explore the label space effectively. We also explore deep neural network for learning the embedding function for the feature space as induced by the embedding label space. Extensive experimental results corroborate the efficacy of our method. We leave for future work for more advanced training mechanism for end-to-end deep learning paradigm for XML classification.

\section{Acknowledgments}
This work is supported by the NSFC (Grant No.11501210, 61602176, 61672231 and 61628203), the NSFC-Zhejiang Joint Fund for the Integration of Industrialization and Information (Grant No. U1609220)
and the Key Program of Shanghai Science and
Technology Commission (Grant No. 15JC1401700). We also would like to thank Liwei Wang for discussions on label graph modeling.

\bibliographystyle{ACM-Reference-Format}
\balance
\bibliography{xml}

\end{document}